\UseRawInputEncoding
\pdfoutput=1

\documentclass[letterpaper, 10 pt, conference]{ieeeconf}  

\IEEEoverridecommandlockouts                              

\usepackage[T1]{fontenc}                        
\usepackage{lmodern}                            
\usepackage[cmex10]{amsmath}                    
\usepackage{amssymb}                            
\usepackage{dsfont}                             
\usepackage{mathrsfs}                           
\usepackage[pdftex]{graphicx}                   
\usepackage{array}                              
\usepackage{upgreek}                            
\usepackage[noadjust]{cite}                     
\usepackage{stfloats}                           
\usepackage{float}
\usepackage{multirow}                           
\usepackage{subfig}                             
\usepackage{fancyhdr}                           
\usepackage[official,left]{eurosym}             
\usepackage{xspace}                             
\usepackage{verbatim}                           
\usepackage[dutch,USenglish]{babel}             
\usepackage{wrapfig}                            
\usepackage{longtable}                          
\usepackage{pgfplots}                           
\usepackage{calc}                               
\usepackage{lscape}								
\usepackage[breaklinks=true,hidelinks,          
            bookmarksnumbered=true]{hyperref}   
\usepackage{vhistory}
\usepackage{makecell}
\usepackage[toc]{glossaries}                    
\usepackage[normalem]{ulem}                     
\usepackage{booktabs}							
\usepackage[bottom]{footmisc}                   
\usepackage{csvsimple}  						
\usepackage{makecell}  						
\usepackage{longtable}
\usepackage{acronym}                            
\usepackage{titlesec}
\usepackage{setspace}
\usepackage{setspace}
\usepackage{microtype}
\usepackage{listings}
\usepackage{xcolor}

\usepackage{algorithm, algorithmic, xpatch}

\usepackage{paralist}                           

\newcommand\copyrighttext{%
  \footnotesize \textcopyright 2021 IEEE. Personal use of this material is permitted.
  Permission from IEEE must be obtained for all other uses, in any current or future 
  media, including reprinting/republishing this material for advertising or promotional 
  purposes, creating new collective works, for resale or redistribution to servers or 
  lists, or reuse of any copyrighted component of this work in other works.  
  }
\newcommand\copyrightnotice{%
\begin{tikzpicture}[remember picture,overlay]
\node[anchor=south,yshift=10pt] at (current page.south) {\fbox{\parbox{\dimexpr\textwidth-\fboxsep-\fboxrule\relax}{\copyrighttext}}};
\end{tikzpicture}%
}
    
\title{\LARGE \bf
Optimal Placement of Roadside Infrastructure Sensors towards Safer Autonomous Vehicle Deployments
}

\author{Roshan Vijay$^{1}$, Jim Cherian$^{1}$, Rachid Riah$^{1}$, Niels de Boer$^{1}$ and Apratim Choudhury$^{2}$
\thanks{This work is supported by Siemens Mobility Pte. Ltd.}%
\thanks{$^{1}$Authors are with Centre of Excellence for Testing \& Research of AVs - NTU (CETRAN),
        Nanyang Technological University, CleanTech One, 637141, Singapore
        {\tt\small \{rvijay,jcherian,rachid.riah,niels.deboer\}@ntu.edu.sg}}%
\thanks{$^{2}$Author is with Siemens Mobility Pte. Ltd., Singapore
        {\tt\small apratim.choudhury@siemens.com}}%
}

\begin{document}

\maketitle
\copyrightnotice
\thispagestyle{empty}
\pagestyle{empty}

\begin{abstract}
	
	Vehicles with driving automation are increasingly being developed for deployment across the world. However, the onboard sensing and perception capabilities of such automated or autonomous vehicles (AV) may not be sufficient to ensure safety under all scenarios and contexts. Infrastructure-augmented environment perception using roadside infrastructure sensors can be considered as an effective solution, at least for selected regions of interest such as urban road intersections or curved roads that present occlusions to the AV. However, they incur significant costs for procurement, installation and maintenance. Therefore these sensors must be placed strategically and optimally to yield maximum benefits in terms of the overall safety of road users. In this paper, we propose a novel methodology towards obtaining an optimal placement of V2X (Vehicle-to-everything) infrastructure sensors, which is  particularly attractive to urban AV deployments, with various considerations including costs, coverage and redundancy. We combine the latest advances made in raycasting and linear optimization literature to deliver a tool for urban city  planners, traffic analysis and AV deployment operators. Through experimental evaluation in representative environments, we prove the benefits and practicality of our approach.
	
\end{abstract}

\section{Introduction}
\label{sec:introduction}

\subsection{Motivation}
\label{ssec:motivation}

Development of autonomous vehicles (AV) has received global attention in the recent years. An AV consists of a vehicle equipped with an Automated Driving System (ADS) \cite{standard2021j3016}. The ADS utilizes various on-board sensors, data processing and compute capabilities to sense and perceive the state of the world around the AV, to plan and to act, in order to \textit{safely} navigate to a given destination.

The sensing and perception capabilities of an AV by means of on-board sensors alone, may be limited and insufficient for urban environments, due to occlusions or limited range, even when their placement on the AV is optimal \cite{liu2019icra}. In contrast, combining the data from an AV's on-board sensors with that received from infrastructure-mounted sensors using Vehicle-to-Infrastructure (V2I) communication is a way to enhance safety and performance. Additional benefits of roadside infrastructure sensors include the possibility that the sensed data can be used to improve traffic coordination and flow, by integrating it with traffic light controller(s).

To accelerate safe AV deployments, infrastructure sensors need to be used in complex urban environments instead of only some controlled testing/trial facilities. This may impose tighter requirements such as complete coverage of intersections with multiple buildings and obstacles in highly built-up areas. Thus, the optimal placement of infrastructure sensors will be a critical factor in enhancing the real-world operational safety of the AV.

\subsection{Problem Statement}
\label{ssec:prob_statement}
The optimal sensor placement problem (OSP) has been extensively researched for surveillance purposes.
In these studies the coverage estimation or the area visible to each sensor is approximated using very simple shapes. For example, in \cite{ahn2016two}, the coverage estimation is approximated using a trapezoid shape, and in \cite{gonzalez2009optimal}, \cite{horster2006approximating}, the area covered by the visibility model of the sensor is described using straightforward relationships. Current studies approach the problem in a theoretical manner, without real-world representations of actual physical sensors or environments.

In order to address these weaknesses, in this paper, we propose a raycasting-based sensor model to approximate accurate line-of-sight coverage. This results in optimal placement even while considering worst case scenarios with multiple 3D obstacles. To solve the optimization problem, we adopt a regularized Binary Integer Programming (BIP) method. We note that BIP is widely used for such discrete sensor placement optimization problems \cite{ahn2016two}.

\subsection{Contributions}
\label{ssec:contributions}
In this paper, we investigate the problem of optimal placement of roadside infrastructure sensors in realistic road networks and make the following contributions:

\begin{enumerate}
    \item A generic Line-of-Sight (LoS) sensor model designed for raycasting and defined using parameters such as vertical and horizontal field of view and range.
    \item A visibility pre-processing methodology that helps to estimate sensor coverage of discrete target points defined within an area of interest.
    \item A framework for optimizing sensor placement to achieve maximum coverage of the given target area while minimizing the number of sensors required.
    \item An end-to-end empirical evaluation of our proposed framework under representative  environments
\end{enumerate}

The rest of this document is structured as follows: in Section~\ref{sec:lit_review} we briefly review the state of the art on sensor placement, coverage calculation and optimization methods. Section~\ref{sec:system_arch} introduces the system architecture of the proposed approach. In Section~\ref{sec:methodology} we describe the theoretical structure of our proposed approach. In Section~\ref{sec:Experimental_evaluation}, we empirically evaluate our proposed methodology and demonstrate its practical usefulness by means of a reference implementation and several input configurations.
We briefly review the potential scope for further improvements and conclude in Section~\ref{sec:Conclusion}.

\section{Literature Review}
\label{sec:lit_review}

The optimal sensors placement problem (OSP) in an outdoor environment is defined similar to the art gallery problem (AGP). AGP is a well-studied visibility problem in computational geometry. It defines the real-world problem of guarding an art gallery with a minimum number of guards who can together observe the whole gallery \cite{urrutia2000art}, \cite{o1987art}. For  outdoor environments, this problem would enquire on how many sensors are necessary to observe an entire Region of Interest (RoI) while considering the presence of obstacles and occlusions. Thus, solving the OSP is similar to solving the AGP \cite{ferreira2010optimizing}, \cite{ahn2016two}. Solving OSP in outdoor environments has received much attention recently, especially to accelerate AV deployments by improving the effective sensor coverage area for a given road network. Several methods have been developed and used in literature to find efficient algorithms to estimate the optimal solution, as the problem is NP-hard \cite{cole1989visibility}.

Research efforts in wireless sensor networks (WSNs) have also studied how to model, simulate and maximize the coverage of a WSN. In \cite{jsan7020020}, Argany et al. proposes an environment framework using a geographic CityGML model of a region which is then converted to either a raster or vector format for estimating sensor coverage. They estimate sensor coverage by quantifying the number of tiles on the raster map which can be `seen' by each sensor. In this approach, the granularity of tile dimensions exerts a strong influence on the performance.

\cite{Akbarzadeh2014} adopts a similar approach by modeling each sensor as a directional probabilistic model, by handling obstacle and topographical occlusions through line-of-sight estimation, and by optimizing the placement of sensors using a gradient descent approach. \cite{geissler2019optimized} involves a grid cell based environment classification approach whereby certain cells are considered to be obstacles or occlusions based on data from a 2D map environment. Using this, 2D sensors are modelled and coverage is estimated for optimizing their placement using a genetic algorithm. \cite{Tan2013} proposes a WSN coverage estimation algorithm based on landmark points, where they act as sampling points for an irregularly shaped RoI. Other studies have also used a similar landmark or target point based approach \cite{chakrabarty2002grid}, \cite{ahn2016two}. \cite{ahn2016two} applies this approach in two phases, first to find the minimum number of sensors using a low resolution grid of target points, and then to maximize coverage by utilizing a higher resolution grid of target points. One common theme among all these studies is that the target point approach seems to be well suited to irregularly shaped RoIs where other regular geometrical coverage estimation methods cannot be utilized.

In terms of optimization algorithms, two different types of OSP formulations have been studied: the MIN and the FIX problems \cite{zhao2013approximate}. The MIN problem aims to minimize the number of sensors needed to cover the entire RoI. The FIX problem aims to maximize the covered area for a \textit{fixed} number of sensors. In this paper, we adopt the MIN approach. The challenges in solving these two problems are well understood and many approximate solutions are proposed in the literature \cite{zhao2013approximate}.

Most studies adopt Binary Integer Programming (BIP) \cite{gonzalez2009optimal}, \cite{aghajan2009multi} as a preferred tool to solve OSP. The advantage of BIP is that the solution of the optimization problem converges to a global optimum and can be used to solve both the FIX and MIN problems depending on formulation. Depending on the coverage estimation methods and the real-world conditions, the optimization algorithms proposed may vary. The commonly used approximation optimization approaches to solve these include genetic algorithms \cite{chen2000camera}, greedy algorithm \cite{aghajan2009multi}, particle swarm optimization \cite{conci2009camera}, \cite{morsly2011particle}, and direct search (DS) method.

In \cite{bianco2012sensor} a raycasting-based approach for area coverage estimation is suggested. They use an optimization algorithm applying the Direct Search (DS) method to solve the FIX problem. 
In our paper, we propose to apply BIP optimization along with raycasting for area coverage estimation to find the optimal number of sensors with their exact positions. Our motivation to use BIP instead of DS is primarily because DS is more well suited to solve the FIX problem, where maximizing the coverage area is less of a priority than fixing the total number of sensors.

BIP offers the following key advantages: 
\begin{inparaenum}[1)] \item The MIN problem can be solved, \item large number of sensors can be considered \end{inparaenum}. On the contrary, the time to converge to an optimal solution can be higher than other methods. Fortunately, this is not a concern for us as our proposed framework is intended to work offline and offsite without any need for instantaneous results.

Our methodology can be of particular interest to practitioners due to its relative simplicity and direct applicability to real-world urban environments and road networks. It requires only a virtual 3D model (`digital twin') of the RoI where the sensors are to be deployed, and it can compute an effective solution while taking real-world dimensions, obstacles, sensor parameters and other constraints into account.

\section{System Architecture}
\label{sec:system_arch}

\begin{figure}[htb]
    \centering
    \centerline{
        \includegraphics[width=\columnwidth]{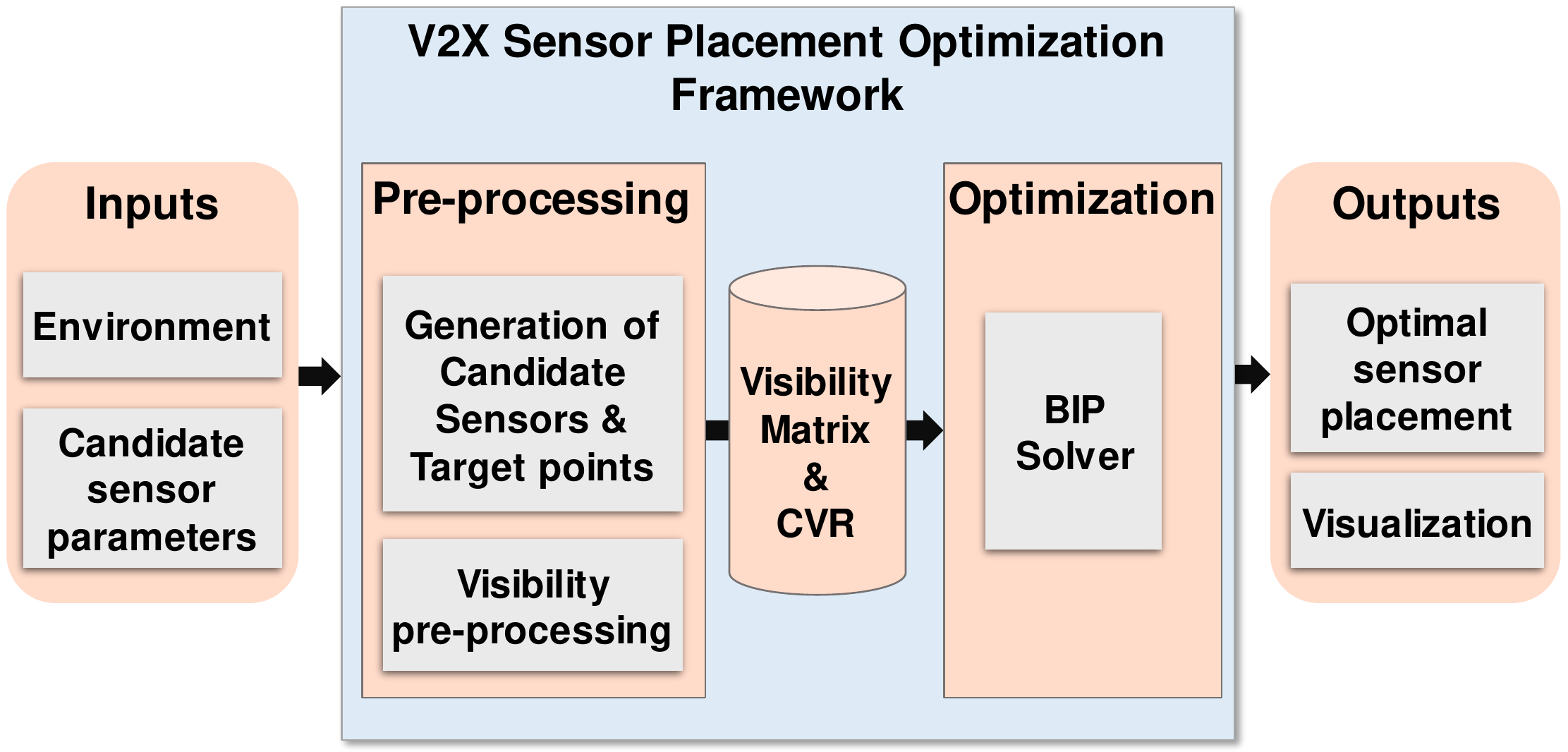}
    }
    \caption{System architecture for the proposed framework}
    \label{fig:framework_architecture}
\end{figure}

The overall system architecture of our proposed approach is illustrated in Fig. \ref{fig:framework_architecture}. 
The inputs are the environment parameters and candidate sensor parameters. The former may include a 3D environment model, road geometry parameters and information about obstacles (if any). The latter forms a superset of `candidate' sensors from which the optimization algorithm can attain the final set of optimal sensor positions. Given these inputs, the framework prepares optimizer inputs required by the BIP solver by generating the candidate sensors, estimating their coverage represented as a visibility matrix produced by the visibility pre-processing step, and calculating the maximum feasible coverage (\textit{CVR}) with an initial superset of candidate sensors. Finally, the solver is invoked and the optimal outputs are generated. The final output is an optimal set of sensor positions along with its visualization.

\section{Methodology}
\label{sec:methodology}
\subsection{Environment model}
\label{ssec:methodology_envModel}
The environment model consists of:
\begin{itemize}
    \item Road network
    \item Road features i.e. lamp posts or sign posts
    \item Obstacles such as buildings and static vehicles
\end{itemize}
\subsection{Sensor model}
\label{ssec:methodology_sensorModel}
In order to estimate sensor coverage, we use raycasting \cite{roth1982ray}, a well known computer graphics technique. It involves sending out virtual rays of light from a diffuse or point source to calculate obstacles in its line-of-sight. In our case, the sensor is modelled as a point source with multiple rays emanating from it. The maximum raycasting distance can be specified as the sensor range.

\begin{figure}[htb]
    \centering
    \centerline{
        \includegraphics[width=0.35\columnwidth]{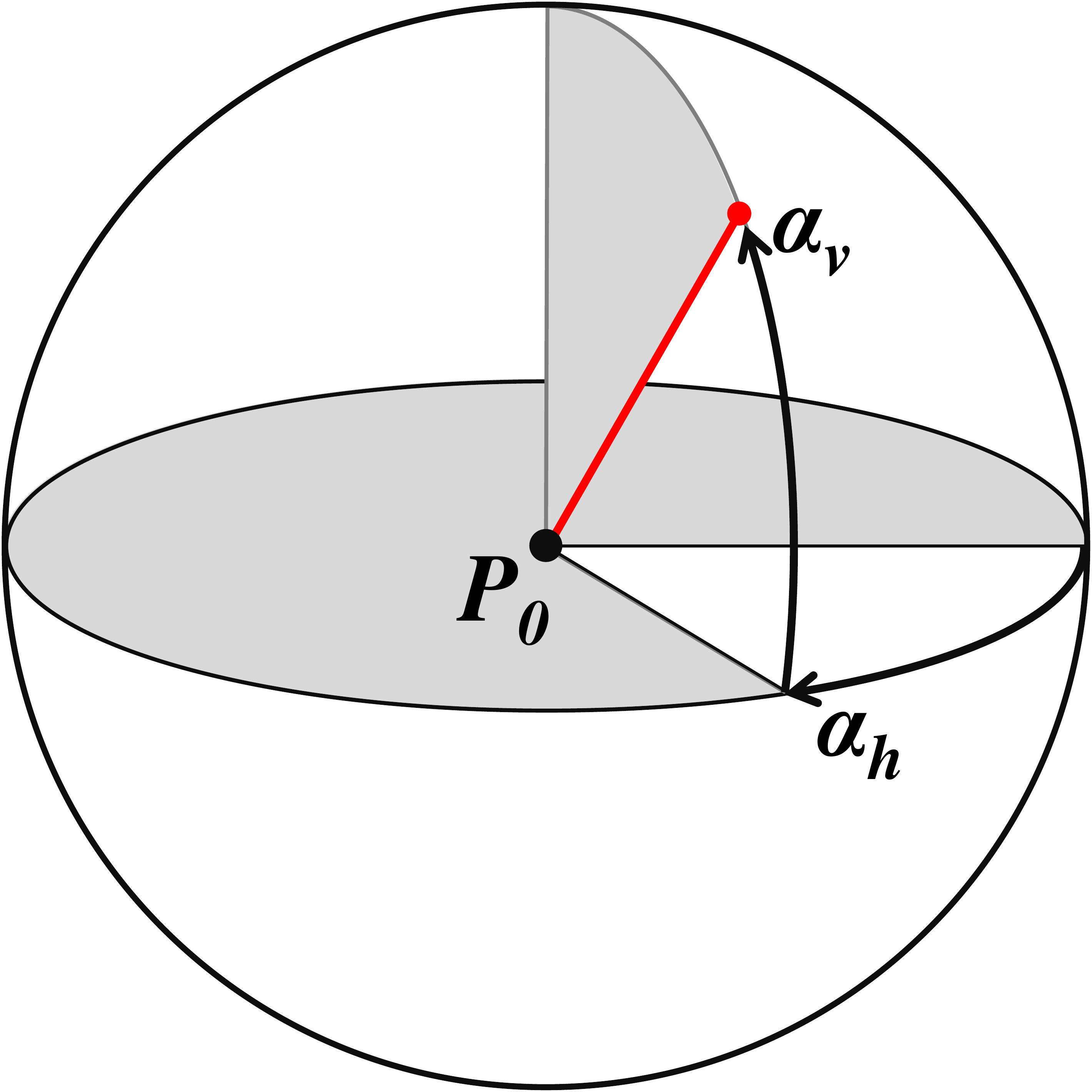}
    }
    \caption{Horizontal coordinate system for ray coordinates}
    \label{fig:Horizontal_coordinates}
    \vspace{-2ex}
\end{figure}

The horizontal coordinate system is used to determine the ray direction vector. This is illustrated in Fig.~\ref{fig:Horizontal_coordinates}.

For a ray emanating from an origin $P_0$, the direction vector $\theta$ of the ray is a function of its horizontal angle $\alpha_h$ and its vertical angle $\alpha_v$:
\begin{equation}
    \theta = [sin(\alpha_h)cos(\alpha_v),\: cos(\alpha_h)cos(\alpha_v),\: sin(\alpha_v)]
\end{equation}
\subsection{Visibility Pre-Processing}
\label{ssec:methodology_vis_preproc}

Once the sensor cast points have been determined, the visibility pre-processing step is necessary to solve the optimization problem. During visibility pre-processing, the list of covered target points is used to construct a binary matrix called the visibility matrix $V$. Its rows and columns represent the candidate sensor locations and covered target points, respectively. A binary value is assigned to each element of $V$ based on if the target point at a particular position is visible to a particular sensor. If the sensor cast point lies within a certain distance to the target point, it is considered visible.

Therefore, an individual element $v_{ik}$ of the visibility matrix $V$ is defined as:
    \begin{equation}
    \label{eq:vis_matrix1}
    v_{ik} = 
        \begin{cases}
            1, & \text{if the target point at position $k$}\\
            &    \text{is covered by the sensor at position $i$}\\
            0, & \text{otherwise}
        \end{cases}
    \end{equation}
    
From the visibility matrix $V$, the maximum feasible coverage \textit{CVR} with all the candidate sensors present is calculated as:
\begin{equation}
\label{eq:CVR_calc}
    \text{\textit{CVR}} = \frac{\text{Number of covered target points}}{\text{Total number of target points}}
\end{equation}

Hence, $\text{\textit{CVR}} \in [0 \quad 1]$\\

The solution of the optimization problem will be a minimum number of sensor positions achieving this \textit{CVR} value. In an ideal case, without occlusions, the maximum feasible \textit{CVR} should be $1$.

If $\text{\textit{CVR}} < 1$ with no obstacles, the number and/or placement of the candidate sensors need to be redefined in order to effectively cover the target region. 

\subsection{Optimization Framework}
\label{ssec:methodology_optimization}
\subsubsection{Assumptions}
\label{sssec:methodology_optimization_assumptions}
The optimization problem is addressed in a deterministic manner where the solution comprising of the minimum number of sensor positions is selected from a set of discrete candidate sensor positions. To ensure that the problem can be applied to real-world scenarios with specific and finite sensor installation locations, a non-deterministic strategy based on randomly selecting candidate sensor positions was not considered.

We assume that the sensors are homogeneous and thus have identical properties such as field-of-view and range and are spaced uniformly within the purview of the permitted candidate sensor locations. We also assume that the target points are defined within an RoI but may not cover the entire RoI and that they are equally spaced and homogenous with respect to the sensor coverage criteria, ie. all target points have the same criteria by which they are considered to be covered by or are `visible' to a sensor.

\begin{figure}[htb]
 	\vspace{-3ex}
	\centering
	\captionsetup{justification=centering}
	\subfloat[]{\label{fig:roadTarget} \includegraphics[width=0.46\columnwidth]{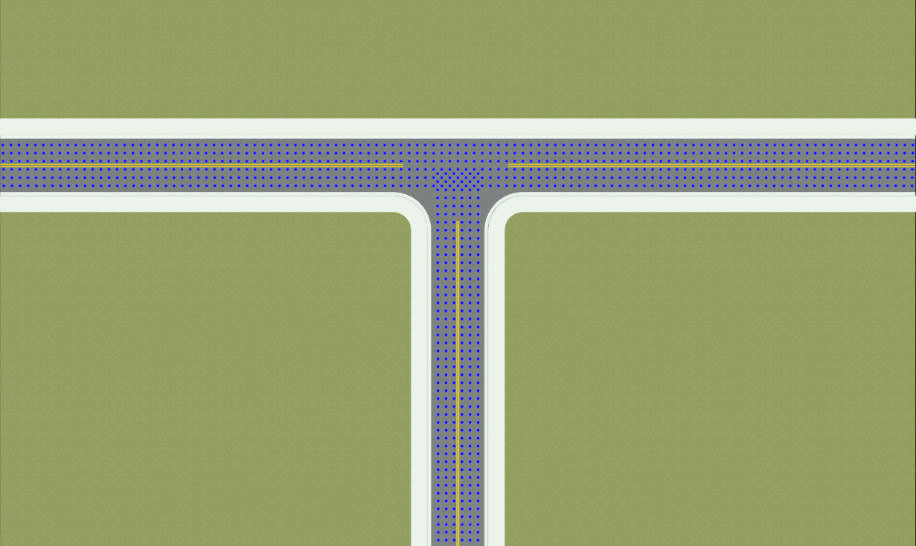}}
	\hspace{1pt}
	\subfloat[]{\label{fig:roadTargetZoomed} \includegraphics[width=0.46\columnwidth]{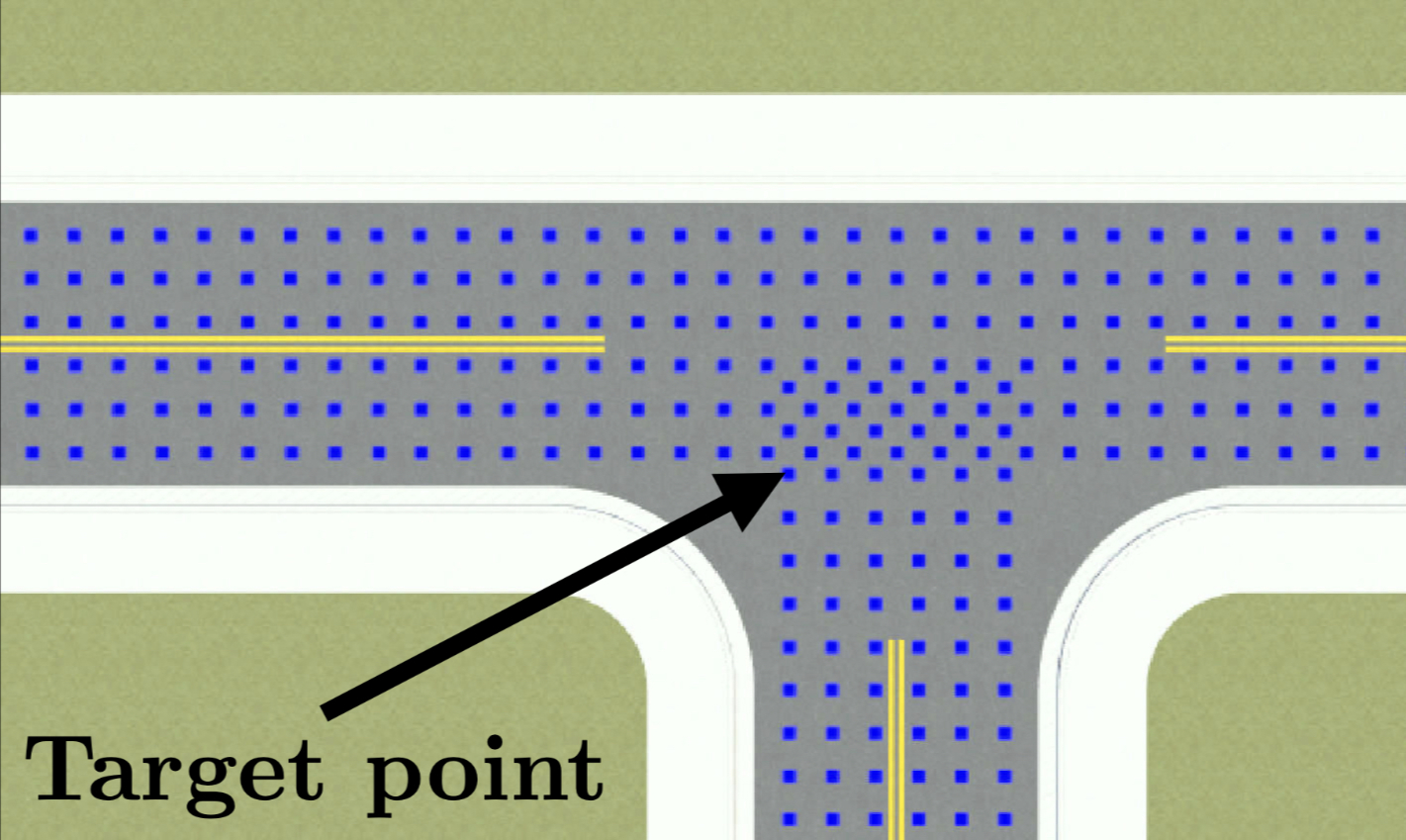}}
	\caption{(a) Illustration of target points (blue dots) on road network, (b) A closer view}
	\label{fig:road_targetPoints}
	\vspace{-1ex}
\end{figure}

The optimization framework is developed and implemented with the following assumptions:
\begin{enumerate}
    \item The RoI consists of the 3D environment, road network and grid of target points located within the road network (see Fig.~\ref{fig:road_targetPoints}). The target points are discrete points in a uniform 2D distribution.
    \item Raycasting is performed across the entire RoI.
    \item A target point is considered visible to the sensor if it is observable from the sensor based on the criteria mentioned in Section~\ref{ssec:methodology_vis_preproc}.
    \item Sensors cannot be placed within the road network. They are defined outside the road boundaries.
    \item All sensors are placed at a specified uniform height $h$ above the ground plane.
\end{enumerate}
\subsubsection{The optimization algorithm}
\label{sssec:methodology_optimization_algo}

The optimization problem is formulated as a BIP problem with binary integer values assigned to the decision variables and is expressed as a linear programming problem. The detailed formulation is given below:
\begin{enumerate}
    \item The variables $N_S$ and $N_T$ are defined as:
    \begin{enumerate}
        \item $N_S =$ number of sensor candidate positions\\
        \item $N_T =$ number of target points\\
    \end{enumerate}
    \item The decision variables $s_i$ and $c_k$ are defined as:
    \begin{equation}
    s_i = 
        \begin{cases}
            1,& \text{if there exists a sensor at position i}\\
            0,              & \text{otherwise}
        \end{cases}
    \end{equation}
    \begin{equation}
    c_k = 
        \begin{cases}
            1,& \text{if the target grid point at position k}\\
              & \text{is covered by atleast one sensor}\\
            0,& \text{otherwise}
        \end{cases}
    \end{equation}
    \item The objective function is defined as:
    \begin{equation}
    \min \quad \left\{ \sum_{i=1}^{N_s} s_i + \lambda \sum_{i=1}^{N_s} D_i \cdot s_i \right\}
    \end{equation}
    
    Where the first term is the main objective function for minimization of sensors and the second is a regularization term. In the regularization term, $D_i$ is a sum of overlap decisions across $s_i$, and $\lambda$ is a calibration parameter. This is introduced to reduce overfitting-like effects such as the selection of very closely spaced sensor positions.
    
    \item $D_i$ is given by:
    \begin{equation}
    D_i = \sum_{j=1}^{N_s} O_{ij}
    \end{equation}
    \begin{equation}
    \label{eq:reg_param_distBinary}
    O_{ij} = 
    \begin{cases}
        1,& \text{if $\| P_j - P_i \| \leq L$}\\
        0,& \text{otherwise}
    \end{cases}
    \end{equation}
    $L$ denotes the desired minimum distance between two candidate sensor positions $P_i$ and $P_j$ for avoiding overlap.\\
    
    \item The first constraint ensures that if the target point $c_k$ is covered then there exists at least one sensor at position $s_i$ to which this target point is visible:
    \begin{equation}
    \label{eq:constraint_1}
        \sum_{i=1}^{N_S} v_{ik} \cdot s_i \geq 0 \quad \text{for} \quad \forall k \quad \in \quad (1, ..., N_T) 
    \end{equation}
    where $v_{ik}$ is defined in Eq.~\ref{eq:vis_matrix1}.
    \item The second constraint ensures that if the target point $c_k$ is covered, then there exists a maximum of $N_S$ sensors to which this target point is visible:
    \begin{equation}
    \label{eq:constraint_2}
        \sum_{i=1}^{N_S} v_{ik} \cdot s_i \leq N_S \cdot c_k \quad \text{for} \quad \forall k \quad \in \quad (1, ..., N_T) 
    \end{equation}
    These constraints are necessary in order to get the value of $c_k$ in Eq.~\ref{eq:constraint_3}.
    \item The third constraint, the coverage constraint, guarantees that at least $N_T \cdot \text{\textit{CVR}}$ target grid points will be covered:
    \begin{equation}
    \label{eq:constraint_3}
    \sum_{k=1}^{N_T} c_k \geq N_T \cdot \text{\textit{CVR}}
    \end{equation}
\end{enumerate}

We note that visibility pre-processing as described in Section~\ref{ssec:methodology_vis_preproc}  helps to make the constraints \ref{eq:constraint_1} and \ref{eq:constraint_2} linear such that the BIP problem can be efficiently solved.

\section{Experimental Evaluation}
\label{sec:Experimental_evaluation}
\subsection{Experimental Setup}
\label{ssec:ExpEval_Exp_Setup}
An experimental setup for the sensor placement optimization framework was implemented using  Python3 scripts. The environment models for raycasting were created using MathWorks RoadRunner \cite{roadrunner} and were imported in the GLTF format \cite{GLTF}. 
Trimesh Python library \cite{trimesh} was used to process the GLTF files and perform raycasting. 
The Python API of lpsolve \cite{lpSolve}, a Mixed Integer Linear Programming (MILP) Solver, was used to solve the BIP problem (refer Section~\ref{sssec:methodology_optimization_algo}).
Finally, the optimal results were visualized using the PyVista library \cite{Sullivan2019pyvista}.

\subsection{Test inputs}
\label{ssec:test_inputs}

\begin{table}[htb]
\centering
\begin{tabular}{|c|c|c|c|}
\hline
Config & \begin{tabular}[c]{@{}c@{}}Candidate\\ spacing\end{tabular} & \begin{tabular}[c]{@{}c@{}}Candidate sensor\\ count ($N_{S}$)\end{tabular} & Obstacles \\ \hline
A1      & 4x4 m                                                       & 79                                                        & 0        \\ \hline
A2      & 4x4 m                                                       & 79                                                        & 4  \\ \hline
B1      & 5x5 m                                                       & 63                                                        & 0        \\ \hline
B2      & 5x5 m                                                       & 63                                                        & 4  \\ \hline
C1      & 6x6 m                                                       & 51                                                        & 0        \\ \hline
C2      & 6x6 m                                                       & 51                                                        & 4  \\ \hline
\end{tabular}
\caption{Test inputs and parameters}
\label{tab:test_inputs}
\end{table}

\begin{figure}[htb]
 	\vspace{-3ex}
	\centering
	\captionsetup{justification=centering}
	\subfloat[]{\label{fig:modelRoad} \includegraphics[width=0.46\columnwidth]{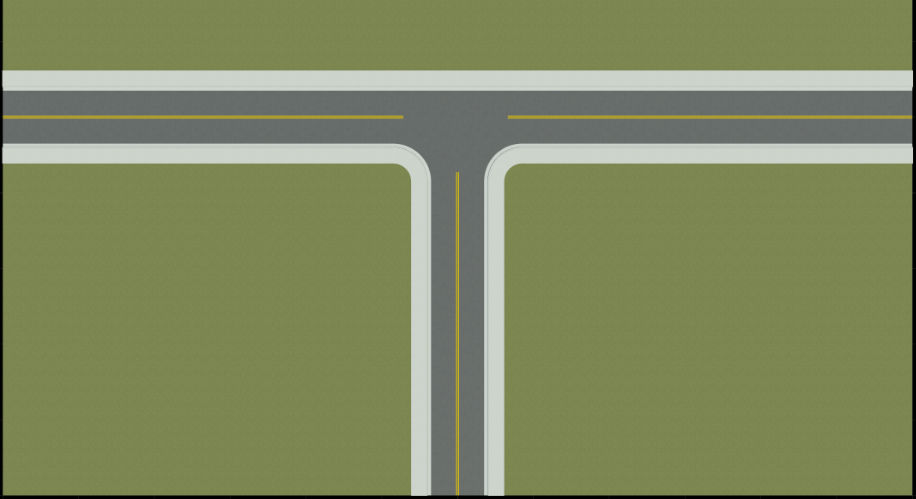}}
	\hspace{1pt}
	\subfloat[]{\label{fig:modelBldg} \includegraphics[width=0.46\columnwidth]{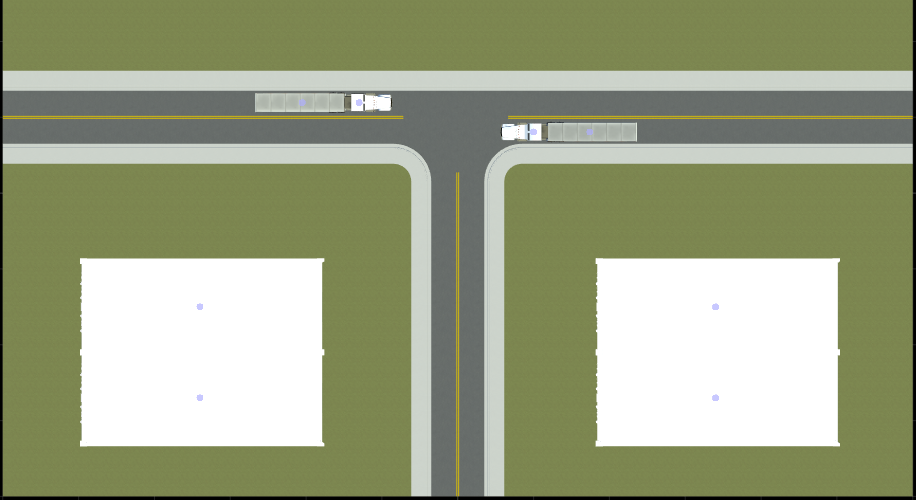}}
	
	\caption{Road geometry environment model: (a) without obstacles (b) with multiple obstacles}
	\label{fig:models_Road_Bldg}
\end{figure}
Six configurations were generated from three candidate sensor arrangement variants, either with or without obstacles. For each configuration, multiple experiments were performed while varying parameters such as sensor height and maximum attainable CVR. They are summarized in  Table~\ref{tab:test_inputs}.
The sensor was modelled with parameters of a commonly used LIDAR sensor 
\cite{mittet2016experimental}, with the following values:
\begin{itemize}
    \item Vertical field of view: $-17^{\circ}$ to $+3^{\circ}$ in $1^{\circ}$ increments
    \item Horizontal field of view: $0^{\circ}$ to $360^{\circ}$ in $1^{\circ}$ increments
    \item Specified range: 100m
    \item Actual range at 2.4m height: $\approx$ 67m
\end{itemize}
Since the sensor model is generic, one may use this to configure other types of LoS sensors as well.

\subsection{Environment setup}
\label{ssec:Eval_EnvSensorModel}

The evaluation was carried out with two environment models shown in Fig. \ref{fig:modelRoad} and \ref{fig:modelBldg}. Both represent a simple T junction with two roads; one with only the 2D road geometry and the other with two sets of oversized containers and two trucks with semi-trailers added as occlusions. The target points were configured such that they cover only the asphalt section of the road.
The target points were configured with the following parameters:
\begin{itemize}
    \item Target point spacing: 1x1m
    \item Target point radius: 1m
    \item Total number of target points: 1008
\end{itemize}

\begin{figure*}[tb]
	\centering
	\captionsetup{justification=centering}
	\subfloat[]{\label{fig:resultA} \includegraphics[width=0.3\textwidth]{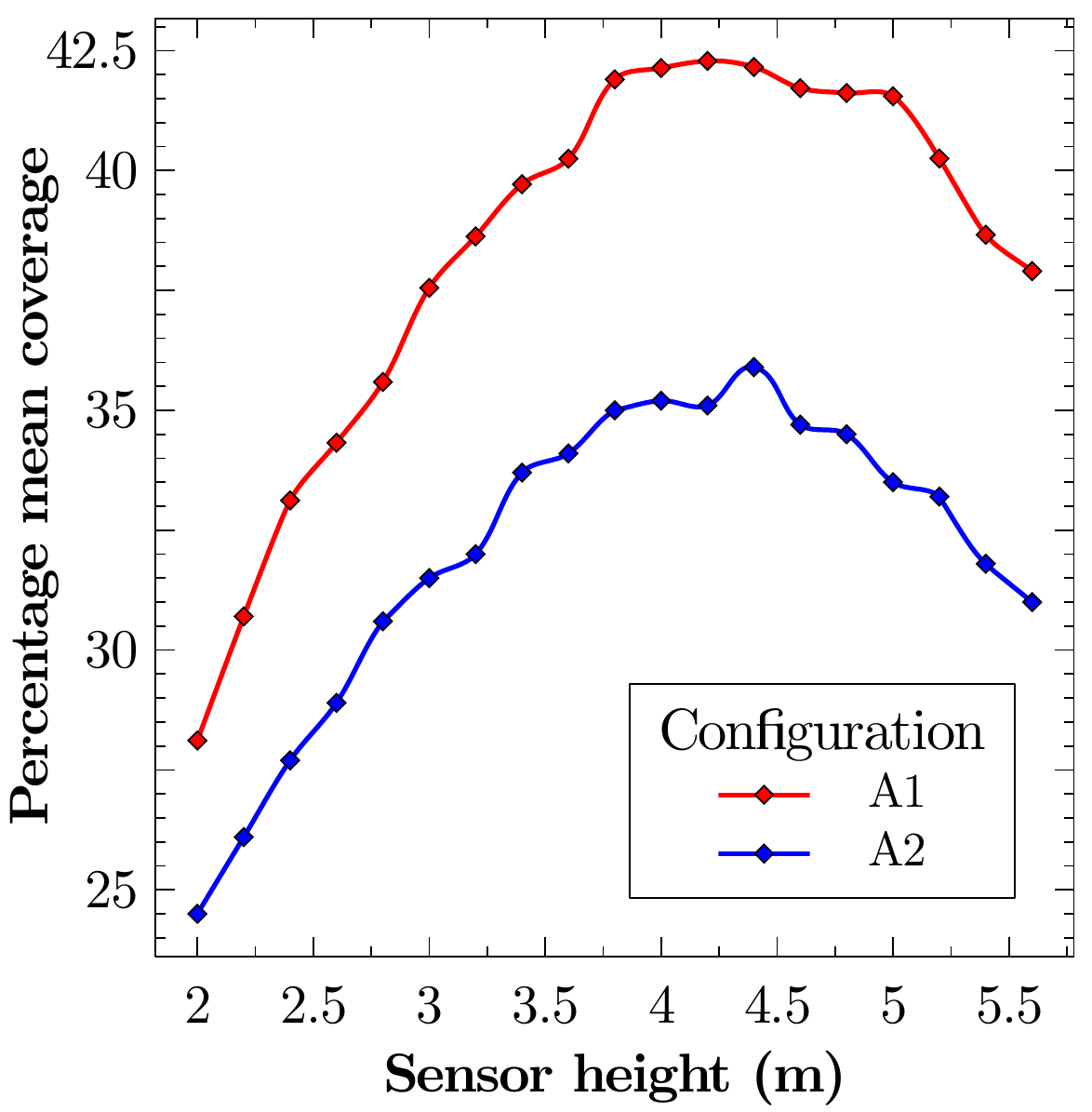}}
	\hspace{5pt}
	\subfloat[]{\label{fig:resultB} \includegraphics[width=0.3\textwidth]{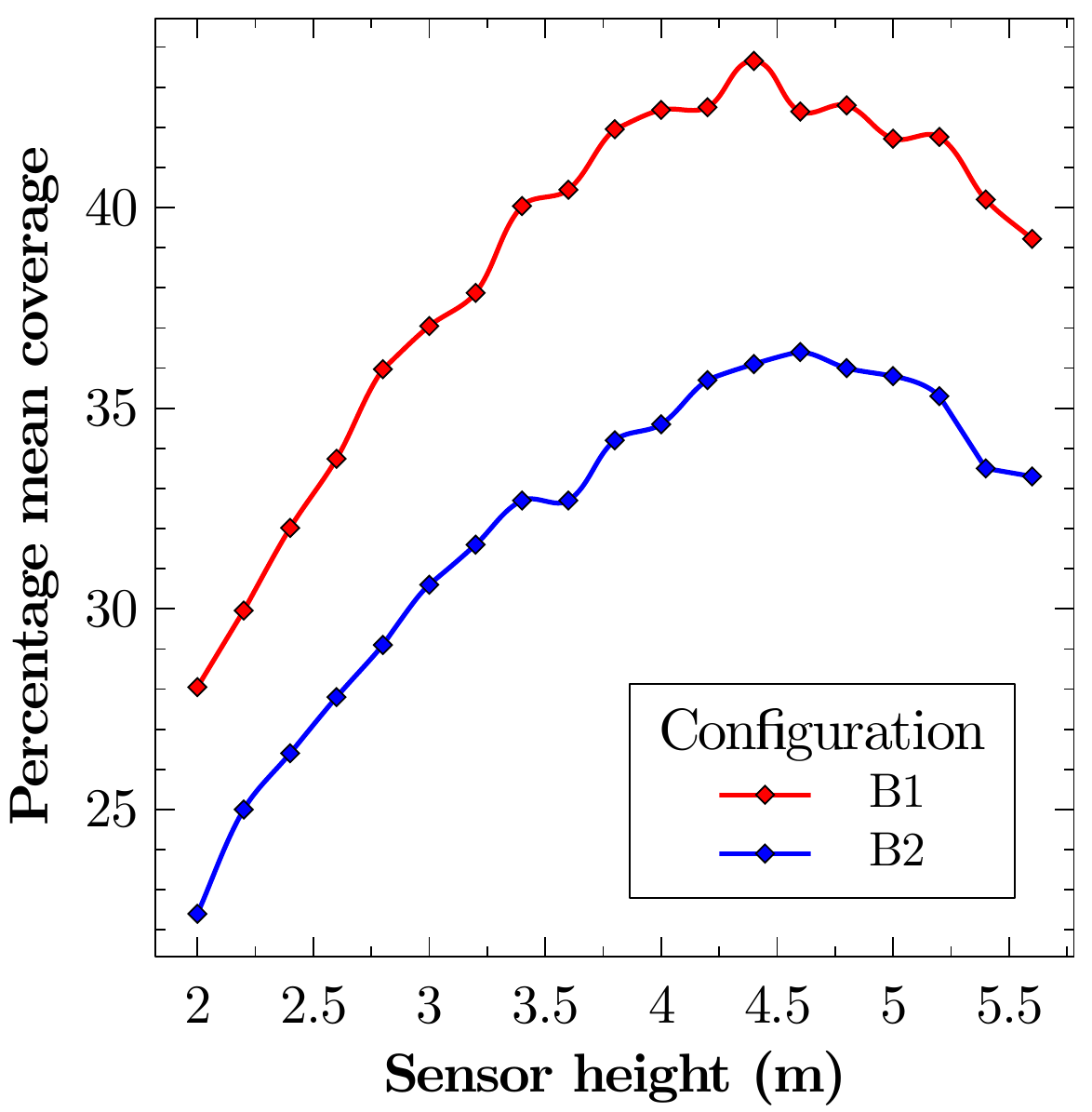}}
	\hspace{5pt}	
	\subfloat[]{\label{fig:resultC} \includegraphics[width=0.3\textwidth]{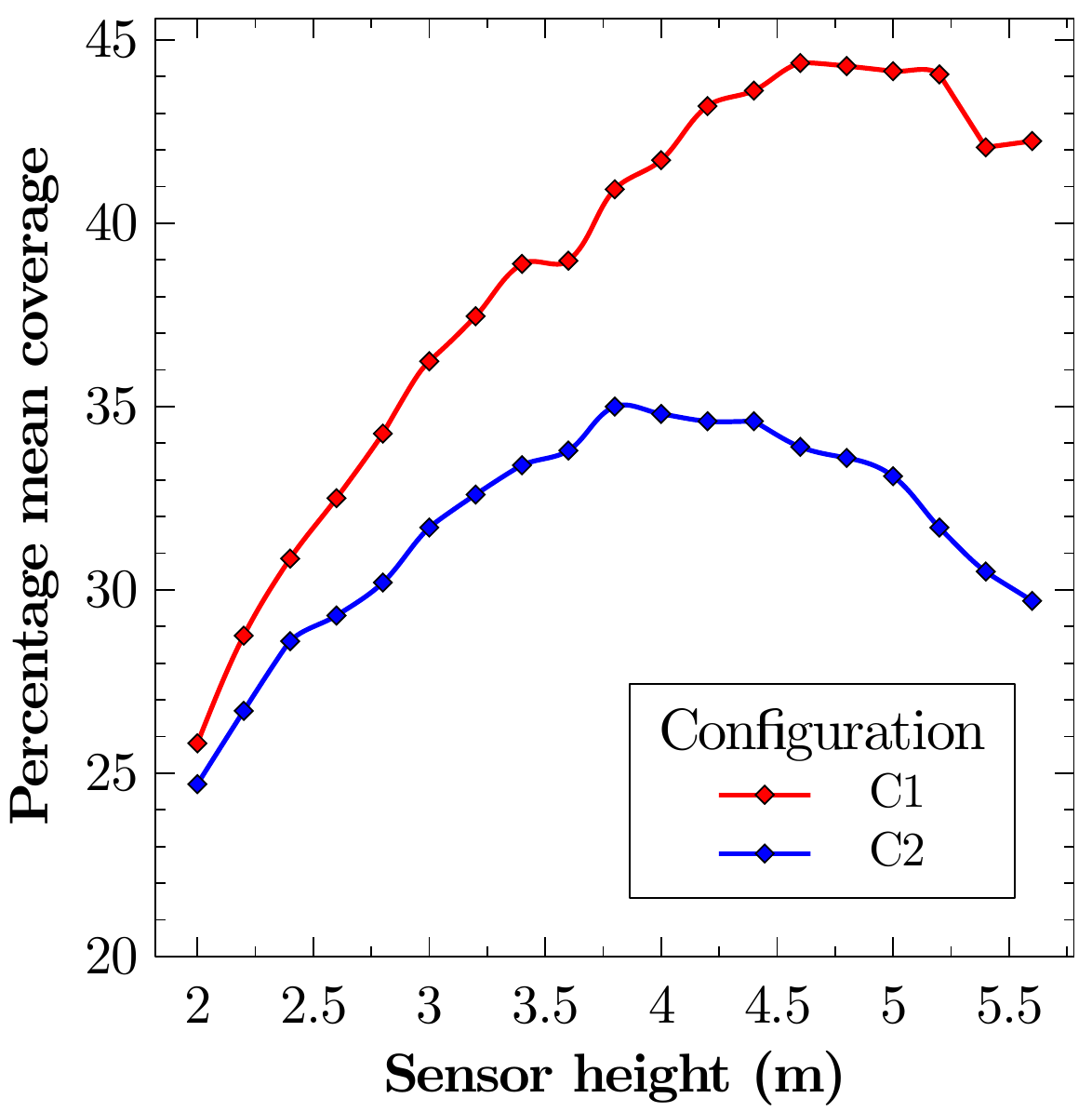}}	
	\caption{Visibility pre-processing coverage performance for sensor configurations:\\(a) A1, A2 (b) B1, B2 and (c) C1, C2, under different sensor heights}
	\vspace{-2ex}
	\label{fig:visResults}
\end{figure*}

\subsection{List of experiments}
\label{ssec:ExpEval_ListofEx}

\subsubsection{Visibility pre-processing performance}
\label{sssec:Eval_visPreProc}
We assess the performance of the visibility pre-processing algorithm by varying parameters relating to the candidate sensor configuration and environment, and evaluating the coverage estimation by calculating the percentage mean coverage of the target points. This refers to the percentage mean count of sensors that can `see' each target point within the RoI. The percentage coverage $C_i$ at the $i$th target point, is calculated as follows:
\begin{equation}
    C_i = \frac{N_{Si}}{N_S} \times 100 
\end{equation}
where $N_{Si}$ is the count of sensors visible at the target point $i$ and $N_S$ is the count of the total number of candidate sensors. We compute the mean of $C_i$ across all the target points to obtain the percentage mean coverage of the selected sensor and configuration. Similarly we employ another metric, namely, the percentage median coverage, to estimate the coverage performance of the selected sensors after optimization.

\begin{figure*}[tb]
	\centering
	\captionsetup{justification=centering}
	\subfloat[]{\label{fig:configA} \includegraphics[width=0.3\textwidth]{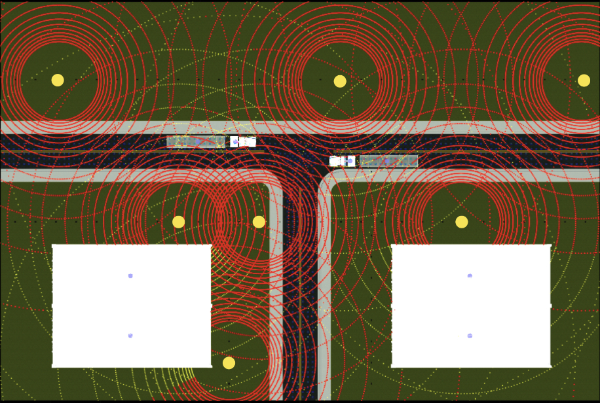}}
	\hspace{5pt}
	\subfloat[]{\label{fig:configB} \includegraphics[width=0.3\textwidth]{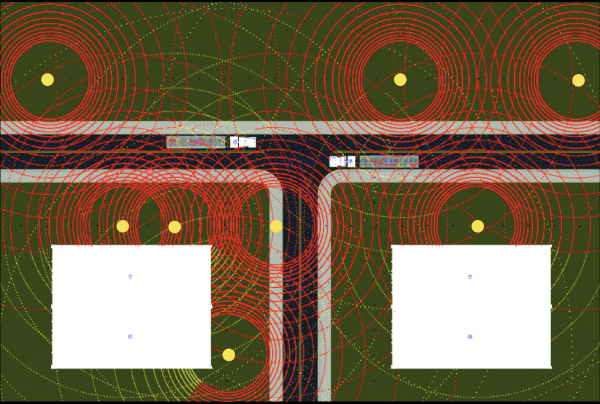}}
	\hspace{5pt}	
	\subfloat[]{\label{fig:configC} \includegraphics[width=0.3\textwidth]{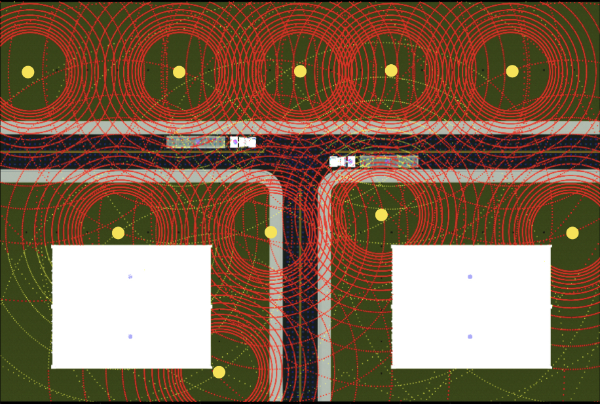}}	
	\caption{Visualization of optimal results for candidate sensor configurations: (a) A2, (b) B2 and (c) C2}
	\label{fig:expt_configs_results_visuals}
\end{figure*}

\begin{figure*}[tb]
\vspace{-1ex}
	\centering
	\captionsetup{justification=centering}
	\subfloat[]{\label{fig:resultD} \includegraphics[width=0.28\textwidth]{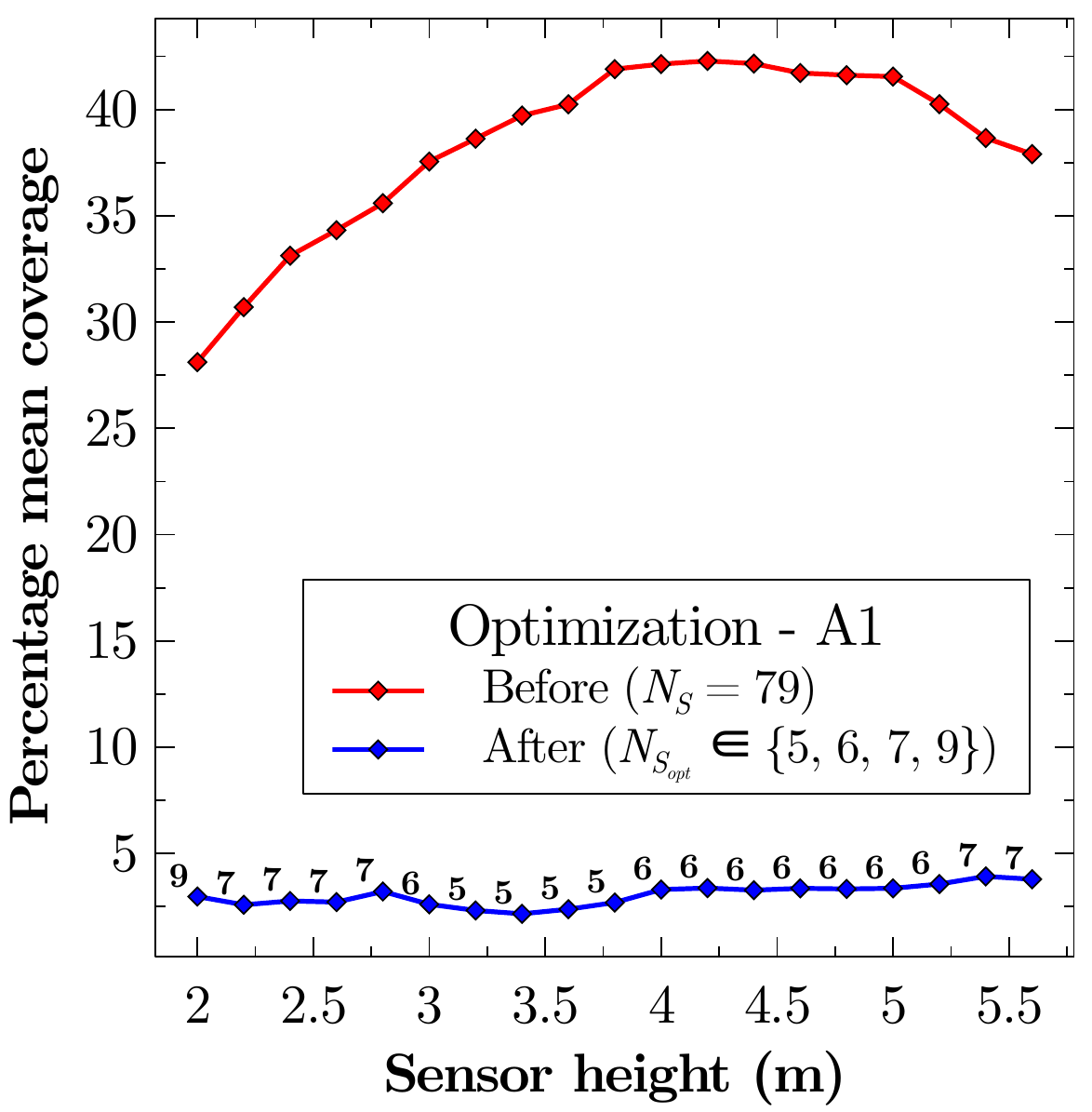}}
	\hspace{5pt}
	\subfloat[]{\label{fig:resultE} \includegraphics[width=0.28\textwidth]{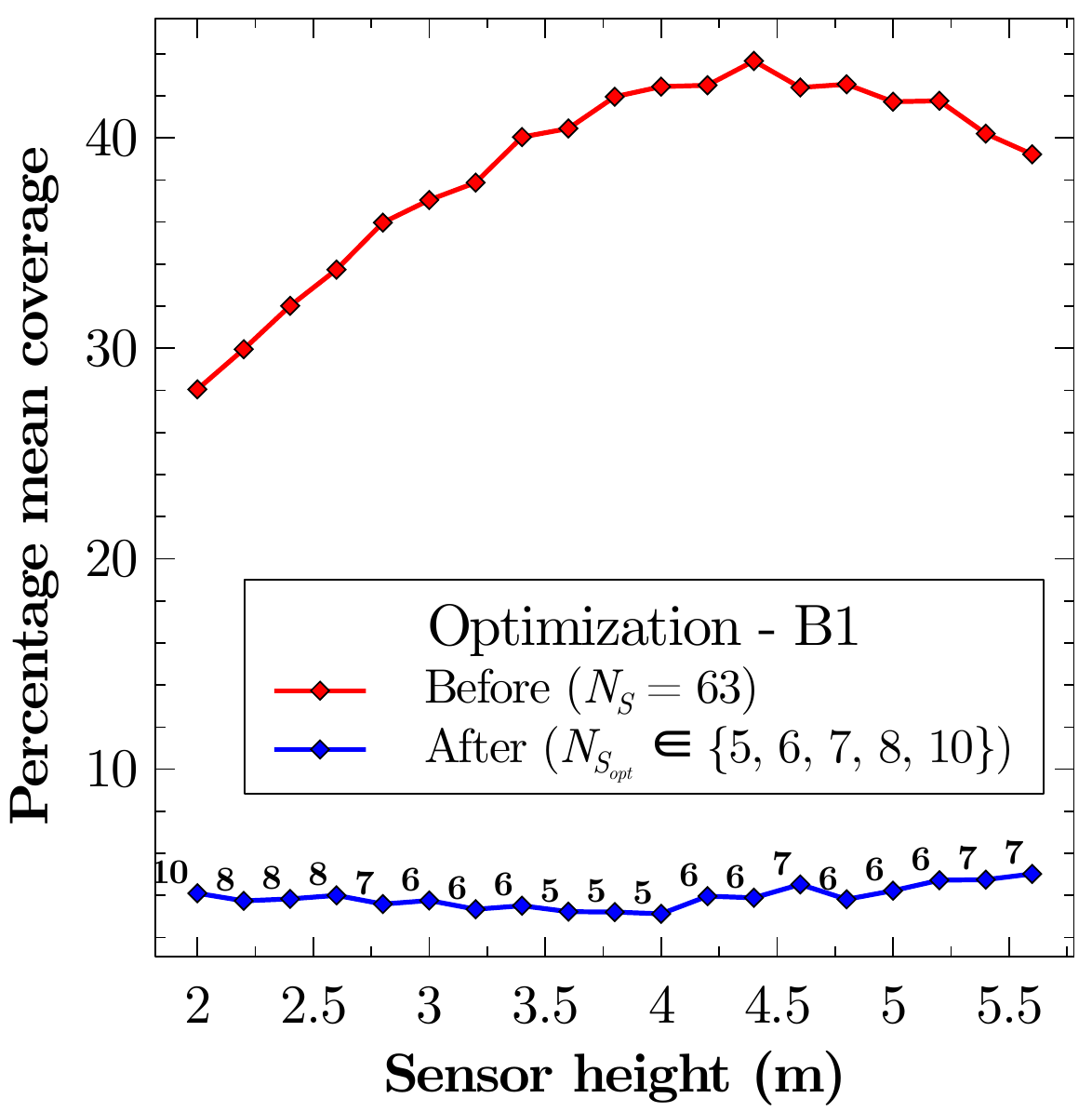}}
	\hspace{5pt}
	\subfloat[]{\label{fig:resultF} \includegraphics[width=0.28\textwidth]{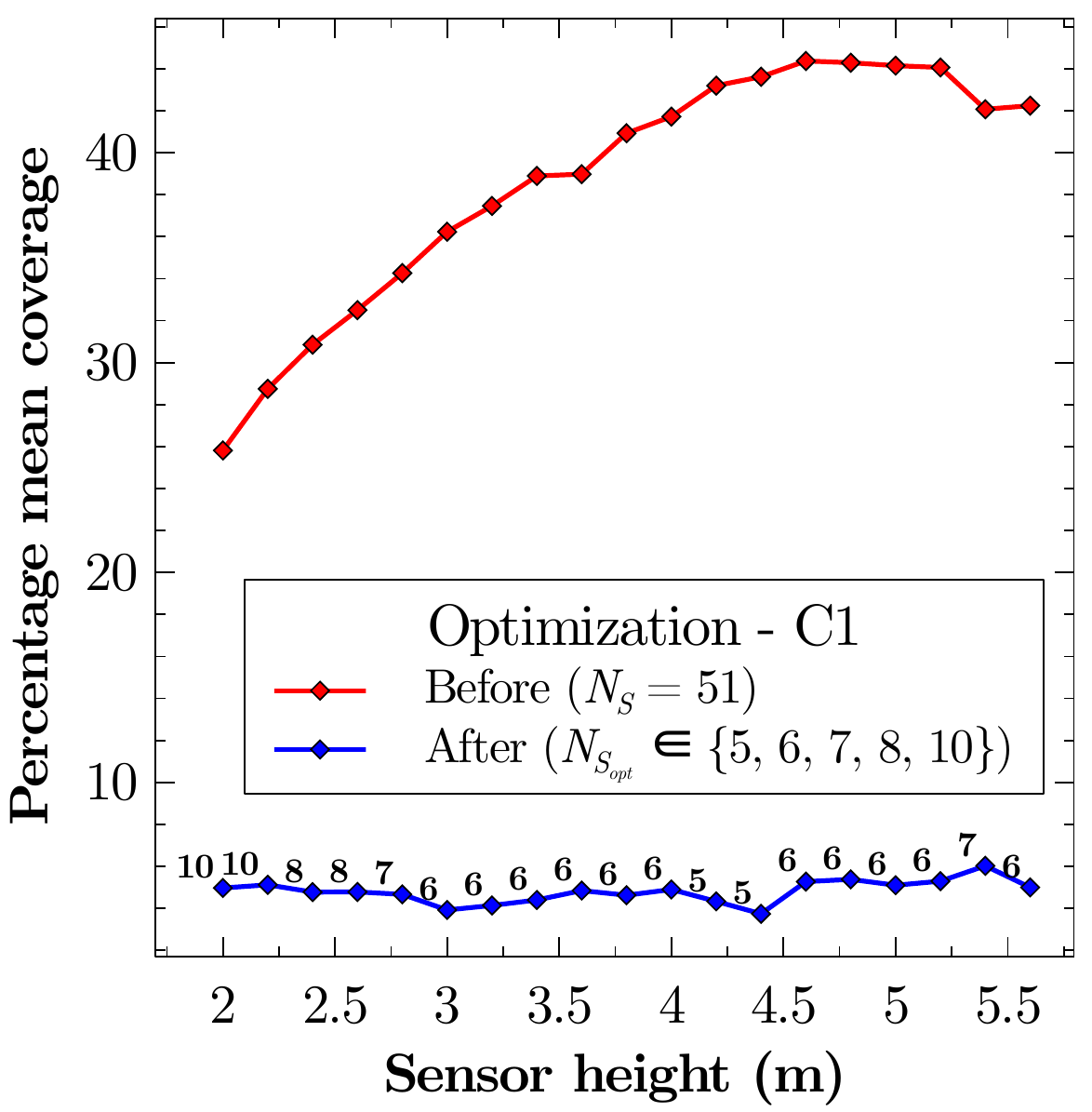}}	
	\caption{
	Performance in terms of coverage of target points before and after optimization},\\ for sensor configurations:(a) A1, (b) B1, (c) C1, under different sensor heights
	\label{fig:optResults}
\end{figure*}

We calculate the percentage mean coverage by:
\begin{itemize}
    \item Varying sensor height
    \item Varying sensor spacing
    \item Varying the presence of obstacles
\end{itemize}
Figs.~\ref{fig:resultA}, \ref{fig:resultB} and \ref{fig:resultC} show the visibility pre-processing performance for the candidate sensor configurations A, B and C respectively with corresponding spacings 4x4m, 5x5m and 6x6m. Separate cases with and without obstacles (A1,A2), (B1,B2) and (C1,C2) are also evaluated. It is evident that the presence of obstacles in the environment greatly reduces the percentage mean coverage performance.
\begin{figure*}[htb]
	\centering
	\vspace{-1ex}
	\captionsetup{justification=centering}
	\subfloat[]{\label{fig:MedianResultA} \includegraphics[width=0.3\textwidth]{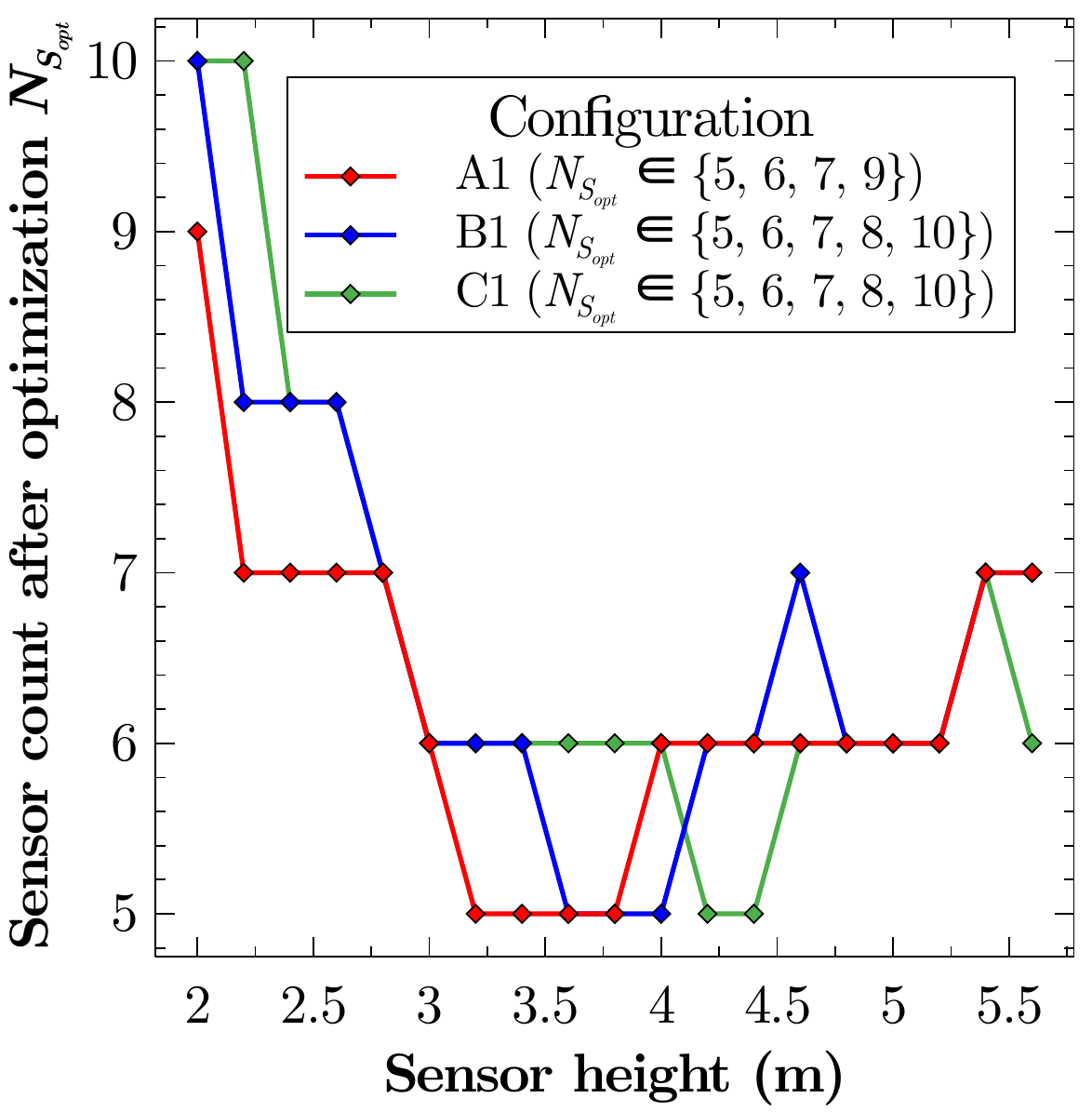}}
	\hspace{5pt}
	\subfloat[]{\label{fig:MedianResultB} \includegraphics[width=0.3\textwidth]{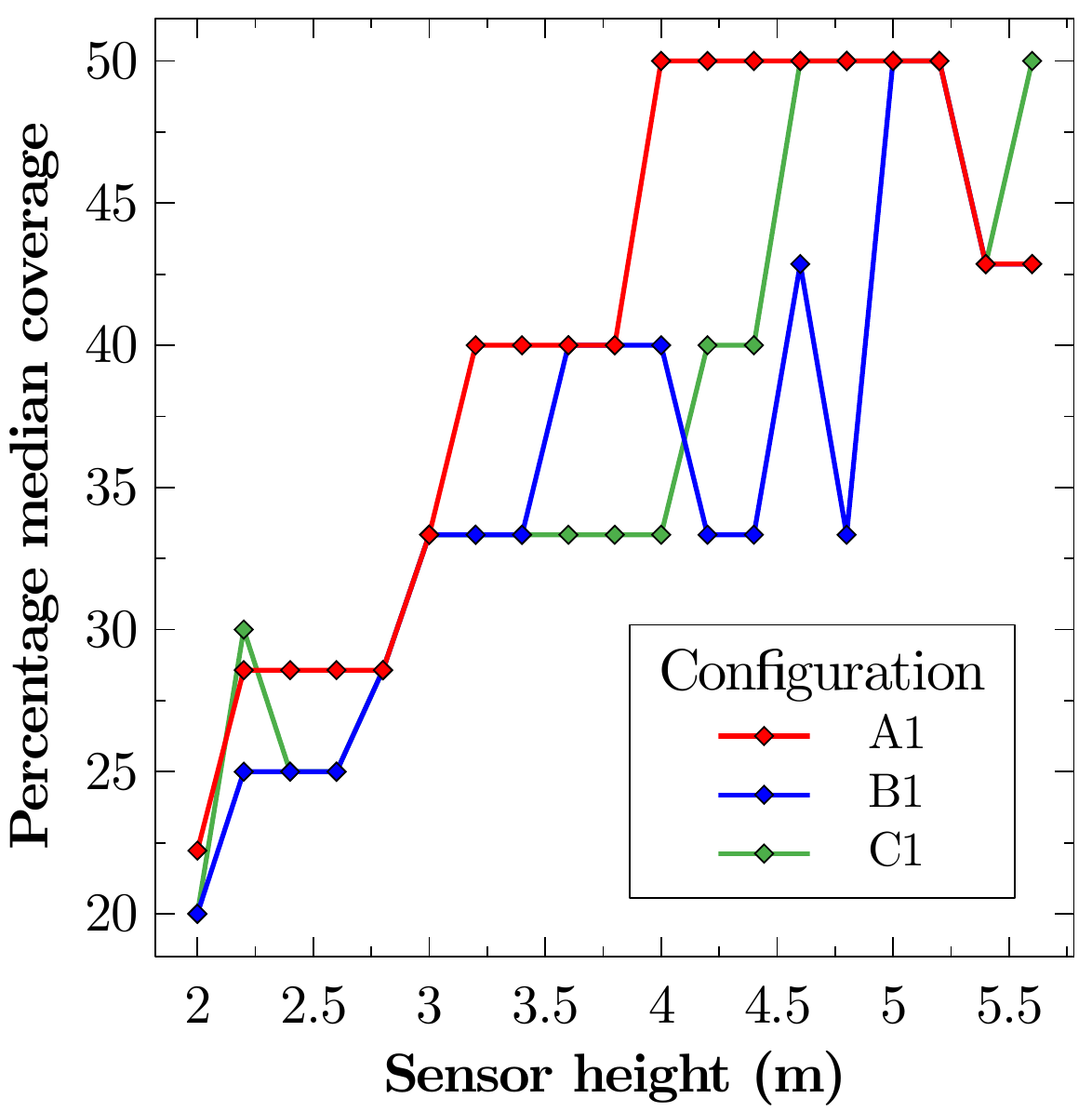}}
	\hspace{5pt}
	\label{fig:vtimeResults}
	\caption{Optimization performance: (a) Number of sensors selected after optimization, $N_{S_{opt}}$\\(b) percentage median coverage after optimization}
	\vspace{-2ex}
\end{figure*}

\subsubsection{Optimization performance}
\label{sssec:Eval_Opt}
To evaluate our proposed optimization approach, we calculate the percentage mean coverage with differing sensor heights for the A2, B2 and C2 configurations as specified in Table~\ref{tab:test_inputs}.

The final optimal sensor configurations for these candidate configurations are visualized in \ref{fig:configA}, \ref{fig:configB} and \ref{fig:configC} respectively. The selected sensor count is significantly lower than the candidate sensor positions. We can also see that closer spaced candidate sensors lead to fewer selected optimal sensors. In Fig.~\ref{fig:configB}, two sensors are found to overlap near the left-side obstacle. This is because our cost function factors the overlapping of sensors as a regularization parameter and not as a hard constraint. We use the percentage mean coverage to compare the initial set of candidate sensors and the final set of selected `optimal' sensors, as shown in Figs.~\ref{fig:resultD}, \ref{fig:resultE} and \ref{fig:resultF} for the configurations A2, B2 and C2 respectively. In order to analyze the optimization performance in a worst case scenario, we choose only the environment model with obstacles. Through this performance comparison, one can discern that that after optimization, the number of redundant sensors covering each target point has greatly reduced; this is because only a fraction of the large set of candidate sensors are selected while maintaining the same \textit{CVR} of 100\%. 
However, we must also consider that even with 100\% coverage, the redundancy at a particular target point may be significantly reduced as the sensor count has been minimized. Fig.~\ref{fig:MedianResultA} shows the percentage median coverage (an indicator of redundancy in terms of sensor coverage) of different configurations after optimization. Clearly, the spacing of the candidate sensors has a direct correlation to the coverage per target point. Closer spaced candidate sensors result in better coverage of each target point after optimization, as the solver has more candidates at its disposal to evaluate and find an optimal subset.

\section{Conclusion}
\label{sec:Conclusion}
In this paper, we have proposed a novel raycasting-based framework for estimating sensor coverage and for optimizing placement of roadside infrastructure sensors using a regularized BIP-based approach.
We have demonstrated the benefits of this framework through experimental results considering different sensor and environment configurations. We note that the framework is able to deliver very useful results when a sufficiently large number of candidate sensor locations are specified.

There are a number of assumptions considered to obtain an optimal solution, such as uniform type and installation height for all sensors and ensuring that the road network is topographically flat and geometrically linear. The coverage estimation is also based on a line-of-sight approach without sensor or environment physics. These are factors which may affect real-world sensor coverage. Therefore, adding the capability of handling curved and irregularly shaped road networks and working with high fidelity physics-based (multi-sensor/environment) simulations, can make the framework more robust.

We also assume that there is no upper limit on the count of optimal sensor locations. This may introduce the need to verify the sensitivity of the performance against the density and distribution of target points and sensors. Future versions of the framework can include an upper limit on maximum permissible sensors after optimization.

\bibliographystyle{IEEEtran}
\bibliography{IEEEabrv,references}

\end{document}